\documentclass[draftclsnofoot, onecolumn]{IEEEtran}
\usepackage[pdftex]{graphicx}
\usepackage{multirow}
\usepackage{bm}
\usepackage[justification=centering]{caption}
\usepackage{cite}
\usepackage{amsmath}
\usepackage{url}

\ifCLASSINFOpdf
\else
\fi
\begin{document}
\title{Mesh Interest Point Detection Based on Geometric Measures and Sparse Refinement}

\author{Xinyu~Lin,
        Ce~Zhu,
        Yipeng~Liu,% <-this % stops a space
\thanks{All the authors are with the School of Information and Communication Engineering/Center for Robotics, University of Electronic Science and Technology
	of China, Chengdu 611731, China.}}

\maketitle

\begin{abstract}
Three dimensional (3D) interest point detection plays a fundamental role in 3D computer vision and graphics. In this paper, we introduce a new method for detecting mesh interest points based on geometric measures and sparse refinement (GMSR). The key point of our approach is to calculate the 3D interest point response function using two intuitive and effective geometric properties of the local surface on a 3D mesh model, namely Euclidean distances between the neighborhood vertices to the tangent plane of a vertex and the angles of normal vectors of them. The response function is defined in multi-scale space and can be utilized to effectively distinguish 3D interest points from edges and flat areas. Those points with local maximal 3D interest point response value are selected as the candidates of 3D interest points. Finally, we utilize an $\ell_0$ norm based optimization method to refine the candidates of 3D interest points by constraining its quality and quantity. Numerical experiments demonstrate that our proposed GMSR based 3D interest point detector outperforms current several state-of-the-art methods for different kinds of 3D mesh models.
\end{abstract}


% \fi
\IEEEpeerreviewmaketitle

\section{Introduction}
\label{introduction}
\IEEEPARstart{W}{ith} the rapid development of 3D image representation and acquisition techniques, more and more 3D image data have been created and widely applied in our daily life. In recent decades, how to effectively process these 3D image data has been one of the hottest topics in the field of 3D computer vision and graphics. As one of the most significant local features in 3D image, 3D interest point can effectively simplify the 3D image data so as to remove the redundancy of 3D image. It lays the foundation of relative computer vision and graphics applications, such as registration \cite{Robust_Global_Registration}, 3D shape retrieval \cite{3D_Salient_spectral_geometric_features,Shape_based_Retrieval_and_Analysis_of_3D_Models}, object recognition \cite{Spin_images}, mesh segmentation \cite{Mesh_segmentation_using_feature_point_and_core_extraction} and simplification \cite{3D_Mesh_saliency_Jour} etc.  

\begin{figure}[tbp]
	\begin{minipage}[b]{1.0\linewidth}
		\centering
		\centerline{\includegraphics[width=0.6\linewidth]{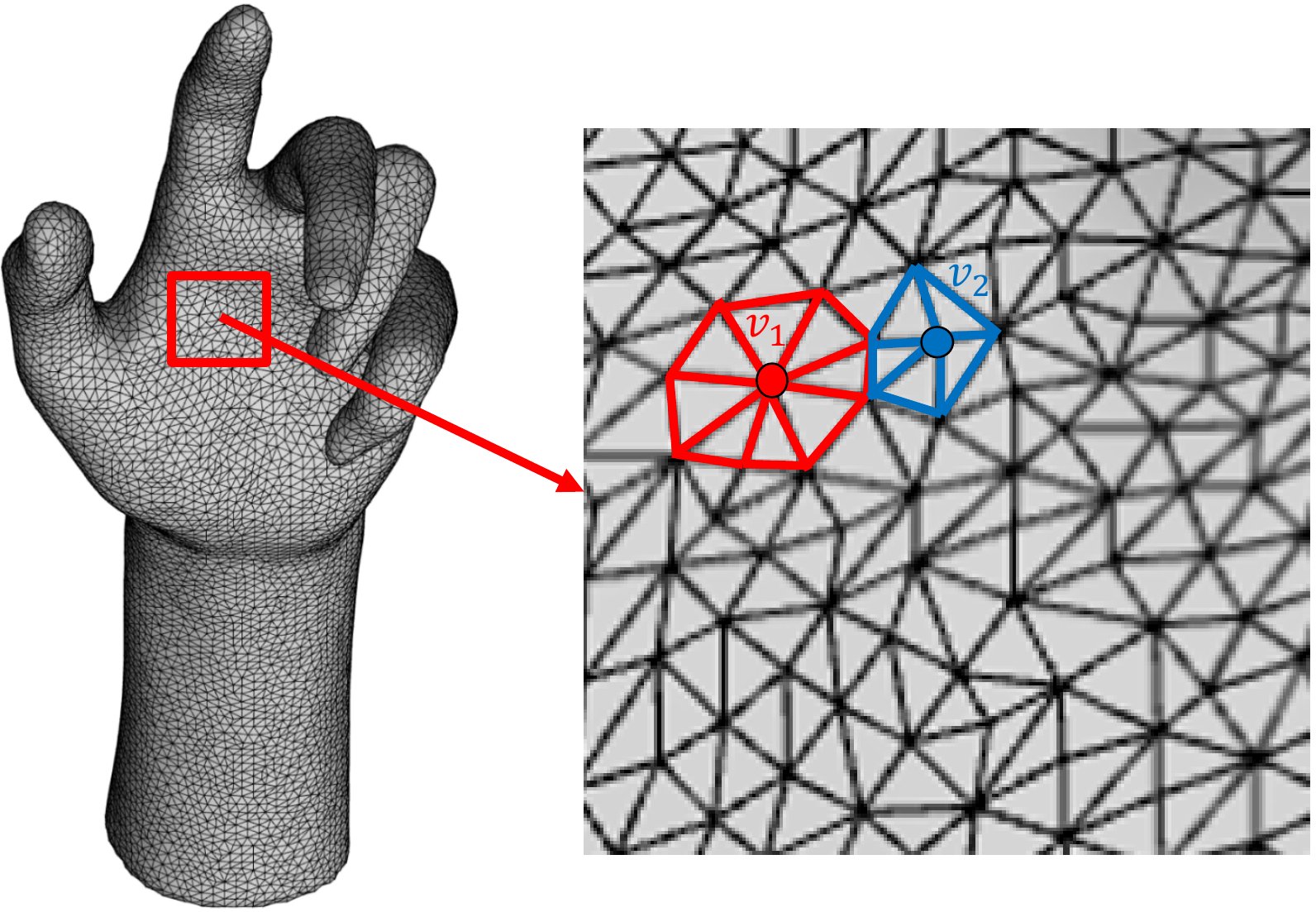}}
	\end{minipage}
	\caption{\emph{hand} 3D mesh model and its local details.}
	\label{3D_mesh_format}
\end{figure}

In this paper, we will focus on the problem of 3D interest point detection for 3D mesh models. Compared with 2D interest point detection algorithms, such as SIFT \cite{SIFT_Jour}, Harris \cite{Harris}, SODC \cite{SODC} and FAST \cite{FAST}, more difficulties exist in 3D interest point detection because of the arbitrary topology of the 3D mesh models. As is shown in Fig. \ref{3D_mesh_format}, a 3D mesh model is represented as a set of vertices and faces  with adjacency information between these entities. For every vertex on the surface of \emph{hand} mesh model, a list of connected vertices is able to formulate triangles with the considered vertex. The topology of 3D mesh model is arbitrary, meaning that every vertex has an arbitrary number of neighborhood vertices. For example, as for the vertices $v_1$ and $v_2$ in Fig. \ref{3D_mesh_format}, the number of neighborhood vertices of $v_1$ is eight while the number of neighborhood vertices of $v_2$ is five, and these distances between neighborhood vertices are totally different. As a consequence, it is challenging to establish a mathematical model to deal with relative issues. Besides, as for 3D mesh models, there is no other useful information except for the spatial position of the vertices and the adjacency information between them. Finally, there is no certain definition for interest point up to now. Different applications may have different requirements for the interest points. For example, some applications are inclined to choose the interest points with high repeatability, some applications prefer those interest points with higher discriminability, and others may require interest points with semantic information, such as Schelling point \cite{3D_schelling_points}. So when designing the interest point detection algorithms, we usually need to define the proper interest point response function according to the practical situation.

Fortunately, researchers have proposed a large amount of 3D interest point detection algorithms to conquer the above-mentioned difficulties in recent years, most of which are based on geometric methods \cite{3D_DoN,3D_HK_Keypoint,3D_Harris_Jour,3D_Cluster_based_point_set_saliency,3D_ISS,3D_KPQ_auto_scale,3D_KPQ_fixed_scale,3D_LSP_and_SI,3D_MeshDoG_Jour,3D_MeshDoG_TIP,3D_Mesh_saliency_Jour,3D_SDTP,3D_SD_corner,3D_SIFT,3D_SIFT_ICIP2015,3D_Salient_point,3D_nSIFT}. Godila et. al. \cite{3D_SIFT} converted the 3D mesh model into voxel grid representation and proposed a method for detecting 3D salient local features inspired by the SIFT algorithm \cite{SIFT_Jour}, in which 3D DoG (Difference of Gaussian) operator is utilized to detect robust 3D interest points. Likely, \cite{3D_MeshDoG_Jour,3D_SIFT_ICIP2015,3D_nSIFT,3D_MeshDoG_TIP} utilized the similar ways to detect 3D interest point. Sipiran and Bustos [6] proposed an effective and efficient extension of the Harris operator \cite{Harris} for 3D objects \cite{3D_Harris_Jour}. Lee et al. \cite{3D_Mesh_saliency_Jour} introduced mesh saliency as a measure of regional importance for 3D meshes, where they defined mesh saliency in a scale-dependent manner using a center-surround operator on Gaussian-weighted mean curvatures. Holte addressed the problem of detecting 3D interest points using Difference-of-Normals operator \cite{3D_DoN}. Castellani et. al. proposed a salient point detection algorithm where sparse 3D keypoints were selected robustly by exploiting visual saliency principles on 3D mesh models \cite{3D_Salient_point}. Except for the geometric methods, there are also some methods detecting 3D interest point in spectrum domain instead of spatial domain \cite{3D_Salient_spectral_geometric_features,3D_HKS,3D_LBSS,3D_Song_mesh_saliency_via_spectral_processing,3D_Song_spectral}. Besides, some researchers formulated 3D interest point detection as a supervised binary classification problem and used learning methods to detect 3D interest point \cite{3D_machine_learning_keypoint,3D_Descriptor_Specific_3D_Keypoint_Detector,3D_RF_keypoint}.

In this paper, based on our previous work \cite{3D_GMSR_Pre}, we propose a new method for detecting 3D interest points of 3D mesh models which are represented by triangular mesh models with arbitrary topology. Two intuitive and effective geometric measures are defined in multi-scale space of 3D mesh models, including the Euclidean distance of neighborhood vertices of vertex $v$ to its tangent plane and the angle of normal vectors between a vertex $v$ and its neighborhood vertices, both of which can be used to distinguish 3D interest points from edges and flat areas effectively. Then we utilize the two geometric measures to calculate the 3D interest point response function for every evolved mesh model in multi-scale space. We utilize the product of these response function value at different scales as the final one, which can effectively improve the 3D saliency value of true 3D interest points and suppress the 3D saliency value of pseudo 3D interest points at the same time. Those vertices with local maxima of 3D saliency value are selected as the candidates of 3D interest points. Finally, we utilize an $\ell_0$ norm based optimization method to refine the 3D interest points from the candidates by constraining the quantity and quality of final 3D interest points. In the hope of spurring further research on this significant topic, we will make our code \footnote{http://www.xylin.net/gmsr/} publicly available.

The rest of this paper is organized as follows. Section II introduces two novel geometric measures of local surface on a 3D mesh model. Section III presents our algorithm - GMSR based 3D interest point detection algorithm. Numerical experiments and results are given in Section VI. Finally, conclusions are drawn in Section V.

\section{Two Geometric Measures}
\label{two_geometric_measures}
\begin{figure*}[tbp]
	\begin{minipage}[b]{1.0\linewidth}
		\centering
		\centerline{\includegraphics[width=\linewidth]{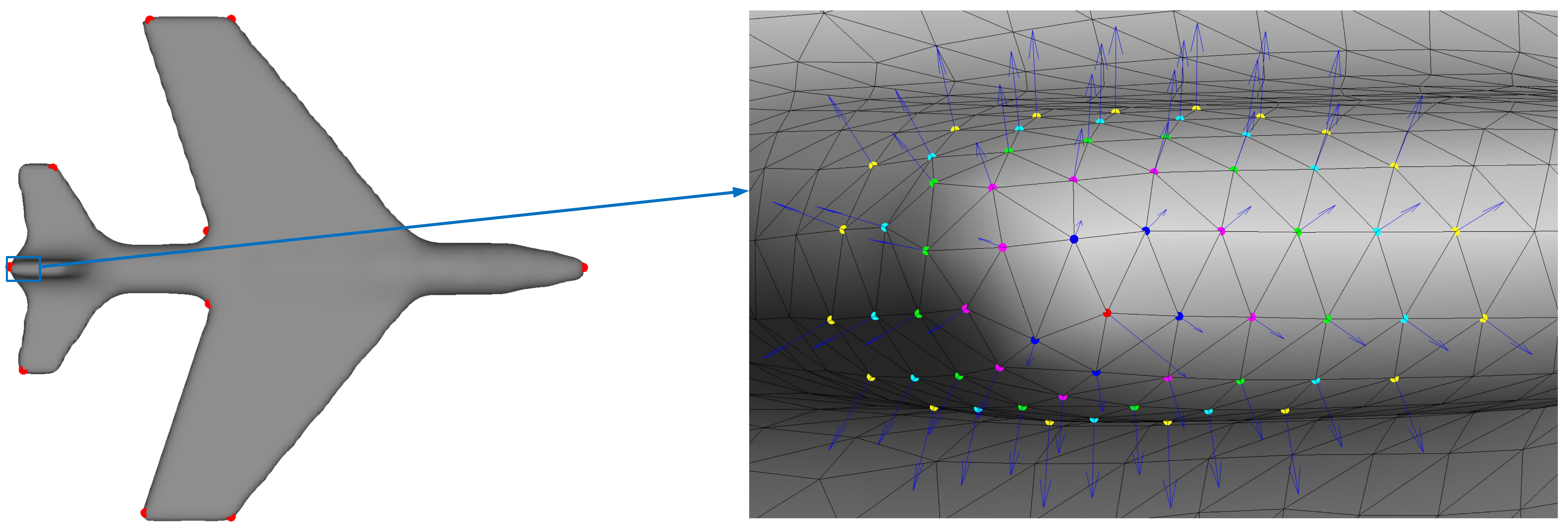}}
	\end{minipage}
	\caption{Left: airplane mesh model; Right: part of airplane mesh model. (Vertex \emph{$v$} (red dot) and its neighborhood rings: \emph{$\bm{V}_1(v)$}, blue dots; \emph{$\bm{V}_2(v)$}, magenta dots; \emph{$\bm{V}_3(v)$}, green dots; \emph{$\bm{V}_4(v)$}, cyan dots; \emph{$\bm{V}_5(v)$}, yellow dots. The arrows represent normal vectors of these vertices.)}
	\label{airplane_detail}
\end{figure*}
According to the second fundamental form of surface \cite{The_second_fundamental_form_of_a_surface}, the most direct measurement which indicates the curvature degree of a vertex on the surface of a 3D mesh model is the Euclidean distances between its neighborhood vertices to the tangent plane of the vertex. Another intuitive measurement is the angles of normal vectors between the vertex and its neighborhood rings \cite{3D_Harris_Jour}. In this paper, we utilize the two simple and intuitive geometric properties together to calculate the interest point response function for every vertex.

\subsection{The Two Geometric Properties}
\label{two_geometric_properties}
As mentioned above, for different vertices on a 3D mesh model, they may have different numbers of neighborhood vertices. Here, we utilize the adaptive technique in \cite{3D_Harris_Jour} to find the neighborhood vertices of a vertex. As shown in Fig. \ref{airplane_detail}, for any vertex $v$ (red dot in Fig. \ref{airplane_detail}) on a triangular mesh model, a list of connected vertices is able to formulate triangles with the vertex, and these connected vertices formulate the first ring neighborhood vertices around $v$, denoted as $\bm{V}_1(v)$ (blue dots in Fig. \ref{airplane_detail}). All the first ring neighborhood vertices of the vertex in $\bm{V}_1(v)$ except for $\bm{V}_1(v)$ and $v$ form the second ring neighborhood vertices around $v$, denoted as $\bm{V}_2(v)$ (magenta dots in Fig. \ref{airplane_detail}). In the same way, we can get the $k$-th ring neighborhood vertices around $v$, denoted as $\bm{V}_k(v),k\in\{1,2,3,...,K\}$. Green dots, cyan dots and yellow dots in Fig. \ref{airplane_detail} represent the second, third and fourth ring neighborhood vertices around \emph{$v$} respectively. The arrows in Fig. \ref{airplane_detail} represent the normal vectors of these vertices. 

For every vertex \emph{$v$} on a 3D mesh model, we utilize \emph{$\bm{n}$} to represent its normal vector. There is only one tangent plane for this vertex \emph{$v$}. The tangent plane can be calculated by 
\begin{equation}
\bm{n}^T[x-x_v,y-y_v,z-z_v]^T=0
\end{equation}
where \emph{$(x_v,y_v,z_v)$} is the coordinate of vertex \emph{$v$}. Let \emph{$\overline{d_k}, 1 \leq k \leq K$} be the harmonic average distance of the \emph{k}-th ring neighborhood vertices to the tangent plane and \emph{$\overline{d}$} be their summation. We can formulate them as: 
\begin{gather}
	\overline{d} = \sum_{k=1}^{K} \overline{d_k} \label{distance_measure}\\
	\overline{d_k} = \frac{W_k}{\sum\limits_{j=1}^{W_k} \frac{1}{d_{kj}}} \label{distance_harmonic}\\
	d_{kj} = \frac{|\bm{n}^T[x_{kj},y_{kj},z_{kj}]^T-\bm{n}^T[x_v,y_v,z_v]^T|}{\lVert \bm{n}\rVert_2}
\end{gather}
where \emph{$(x_{kj},y_{kj},z_{kj})$} is the coordinate of the \emph{j}-th point in \emph{$\bm{V}_k(v)$} and \emph{$W_k$} is the total number of vertices in \emph{$\bm{V}_k(v)$}. 

Let $\bm{n}_{kj}$ be the normal vector of neighborhood vertex $(x_{kj},y_{kj},z_{kj})$.  Let \emph{$\overline{\theta_k}, 1 \leq k \leq K$} be the harmonic average angle between the vertex $v$ and its \emph{k}-th ring neighborhood vertices. \emph{$\overline{\theta}$} be their summation. Similarly, We can formulate them as:
\begin{gather}
	\overline{\theta} = \sum_{k=1}^{K} \overline{\theta_k} \label{theta_measure}\\
	\overline{\theta_k} = \frac{W_k}{\sum\limits_{j=1}^{W_k} \frac{1}{\theta_{kj}}} \label{theta_harmonic}\\
	\theta_{kj} = \text{arccos}(\frac{\bm{n}^T\bm{n}_{kj}}{||\bm{n}||_2||\bm{n}_{kj}||_2})
\end{gather}

\subsection{Analysis}
\begin{figure*}[tbp]
	\begin{minipage}[b]{1.0\linewidth}
		\centering
		\centerline{\includegraphics[width=\linewidth]{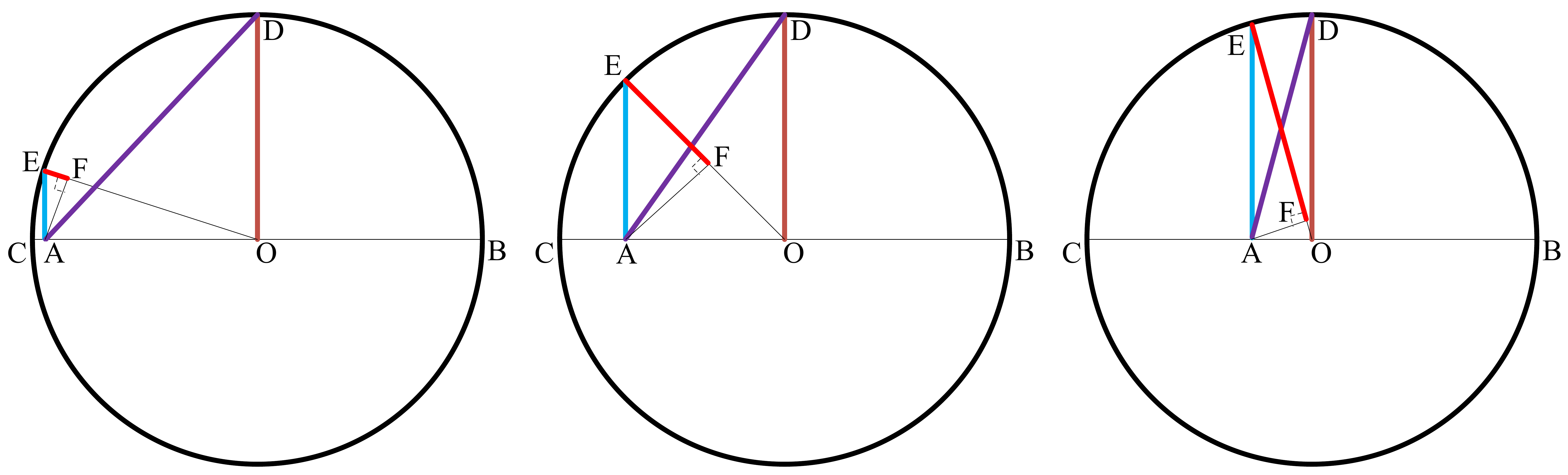}}
	\end{minipage}
	\caption{Four types of mean value of line AB and AC. (EF: harmonic mean; AE: geometric mean; AD: quadratic mean; OD: arithmetic mean.)}
	\label{harmonic_average}
\end{figure*}
Because an arbitrary number of neighborhood vertices exist around vertex \emph{$v$}, the average value $d_{kj}$ and $\theta_{kj}$ are more suitable than accumulative value or multiplied value to measure the regional importance for mesh models. As shown in Fig. \ref{harmonic_average}, under the condition that the total length of line AB and AC is consistent, the harmonic mean of AB and AC, denoted as EF in Fig. \ref{harmonic_average}, is most sensitive to the difference between AB and AC. Compared with other three types of average values, harmonic mean is the most sensitive to the outlier, which could distinguish the flat areas, edges and interest points on a 3D mesh model very well. As shown in TABLE \ref{average_tables}, for flat areas, all the $d_{kj},1\leq k \leq K,1\leq j \leq W_j$ and $\theta_{kj},1\leq k \leq K,1\leq j \leq W_j$ should be small, so the four types of average value of $d_{kj}$ and $\theta_{kj}$ are small too. For all the interest points, all the $d_{kj}$ and $\theta_{kj}$ are large. So the four types of average value of them are also large. But for edges in 3D mesh models, the quadratic mean and arithmetic mean of $d_{kj}$ and $\theta_{kj}$ may be large. The harmonic mean is still small. So, the harmonic average values of $d_{kj}$ and $\theta_{kj}$ in (\ref{distance_harmonic}) and (\ref{theta_harmonic}) could be effectively utilized to detect interest points. To fully utilize the local information around the vertex $v$, we accumulate several harmonic values of $d_{kj}$ and $\theta_{kj}$ together.

\begin{table}[tbp]  
	\centering
	\renewcommand\arraystretch{1.2}
	\caption{The effect of four types of average values of $d_{kj}$ and $\theta_{kj}$ presented on different regions of a 3D mesh model.}
	\begin{tabular}{ccccc}  
		\hline
		\cline{1-5}
		\ &harmonic \ \ &geometric \ \ &arithmetic \ \ &quadratic \\
		\hline
		Flat areas &small &small &small &small \\        
		Edges &small &relatively small &large &large \\        
		Keypoints &large &large &large &large \\        
		\hline
		\cline{1-5}
	\end{tabular}
	\label{average_tables}
\end{table}

\section{GMSR Based 3D Interest Point Detector}
In this section, we describe our proposed GMSR based 3D interest point detection algorithm in detail. The outline of our approach is as follows:
\begin{itemize}
	\item Utilize 3D Gaussian filters to formulate the multi-scale representation for a 3D mesh model \emph{$\bm{M}(x,y,z)$} and obtain a series of evolved 3D mesh models \emph{$\bm{M}_\delta(x,y,z)$}.
	
	\item Define and calculate the 3D interest point response function based on the two geometric measures described in Section \ref{two_geometric_measures} in multi-scale space.
	
	\item Select these vertices with local maxima of final response function value as the candidates of 3D interest points.
	
	\item Utilize an $\ell_0$ norm based optimization method to refine the candidates of 3D interest points. 
\end{itemize}

\subsection{Construction Scale Space}
Similar to the algorithms in\cite{3D_SIFT,3D_Salient_point,3D_Mesh_saliency_Jour}, we utilize the 3D Gaussian filter to construct the multi-scale space for a 3D mesh model. Let \emph{$\bm{M}(x,y,z)$} be a 3D mesh model and \emph{$\bm{M}_\delta(x,y,z)$} be its corresponding evolved 3D mesh modes in the multi-scale space. We can have the following relationship:
\begin{gather}
	\bm{M}_\delta(x,y,z)=\bm{M}(x,y,z) * \bm{G}(x,y,z,\delta), \\
	\bm{G}(x,y,z,\delta) = \frac{1}{(\sqrt{2\pi}\delta)^{3}}e^{-\frac{(x^{2}+y^{2}+z^{2})}{2\delta^2}},
\end{gather}
where \emph{$\delta \in$} \{\emph{$\varepsilon$}, 2\emph{$\varepsilon$}, 3\emph{$\varepsilon$}, ...\} is the standard deviation of 3D Gaussian filter and \emph{$\varepsilon$} amounts to 0.3\% of the length of the main diagonal located in the bounding box of the model \cite{3D_Mesh_saliency_Jour}. $*$ is convolution operator.

\subsection{3D Interest Point Response Function}
\begin{figure}[tbp]
	\begin{minipage}[b]{1.0\linewidth}
		\centering
		\centerline{\includegraphics[width=0.6\linewidth]{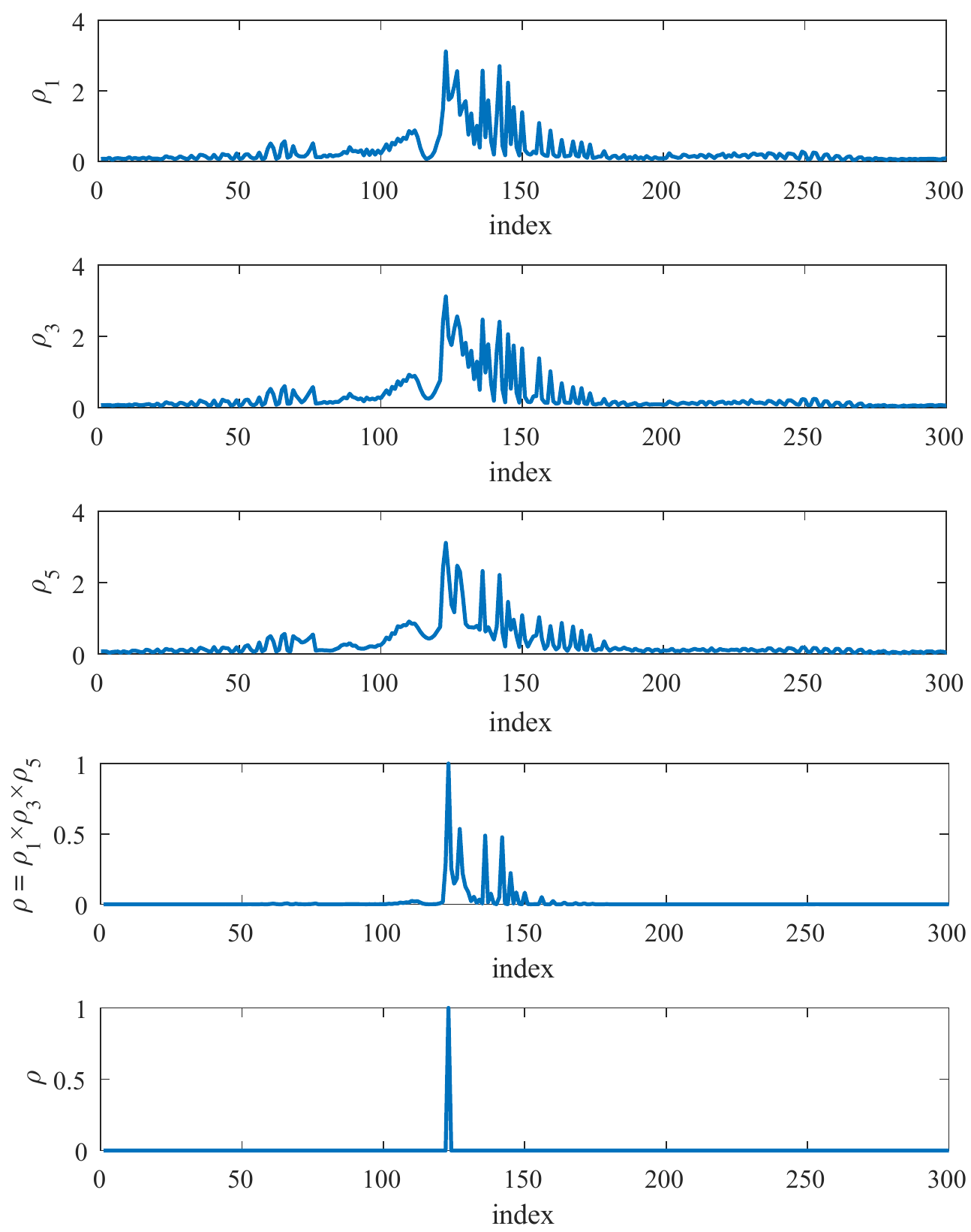}}
	\end{minipage}
	\caption{A part of response function curves of \emph{bird} mesh model. From top to bottom, they represent response function curves at scale $\varepsilon$ ($\rho_1$), scale $3\varepsilon$ ($\rho_3$), scale $5\varepsilon$ ($\rho_5$), final response function curve $\rho=\rho_1 \times \rho_3 \times \rho_5$ and final response function curve after non-maxima suppression respectively.}
	\label{response_curves}
\end{figure}

Based on the two geometric measures of local surface mentioned in Section \ref{two_geometric_measures}, we define a 3D interest point response function \emph{$\rho$}, which enables us to extract perceptually meaningful points of interest from 3D mesh models. Similar to the multi-scale curvature product used in MSCP corner detector \cite{MSCP}, for any vertex $v$ in a 3D mesh model, we utilize the product of response function values at different scales as the final one. So, for any vertex $v$, its final interest point response value $\rho$ can be formulated as: 
\begin{gather}
	\rho = \prod_{s}\rho_s, \\
	\rho_s = \frac{\overline{d_s} - \text{min}(\overline{\bm{d}_s})}{\text{max}(\overline{\bm{d}_s})-\text{min}(\overline{\bm{d}_s})} + \alpha \frac{\overline{\theta_s} - \text{min}(\overline{\bm{\theta}_s})}{\text{max}(\overline{\bm{\theta_s}})-\text{min}(\overline{\bm{\theta}_s})} \label{linera_norm},
\end{gather}
where $\overline{d_s}$ and \emph{$\overline{\theta_s}$} represent the two geometric measures of vertex $v$ in scale $s\varepsilon, s=1,2,...$ mentioned in Section \ref{two_geometric_measures}, and can be obtained via (\ref{distance_measure}) and (\ref{theta_measure}) respectively. The $\overline{d_s}$ of all the vertices on a 3D mesh model in scale $s\varepsilon$ together formulate the $\overline{\bm{d}_s}$, and it is the same to the $\overline{\bm{\theta}_s}$. The first term and the second term of (\ref{linera_norm}) are linear normalization. We combine the two measures together via a linear function to represent their relationship approximately. \emph{$\alpha$} is a weighting factor for two kinds of measures. 
Response function values at different scales $\rho_s$ are multiplied to formulate the final response function value $\rho$. Fig. \ref{response_curves} displays a part of response function curves of \emph{bird} 3D mesh model. We can see that using the multi-scale response function value product could effectively improve the 3D interest point response function value \emph{$\rho$} of true 3D interest points and suppress the value \emph{$\rho$} of pseudo 3D interest points. The final response function map of \emph{bird} mesh model can be found in the left part of Fig. \ref{GMSR_bird_2}.

\subsection{Selecting the Candidates of 3D Interest Points}
\label{non-maxima_supression}
After we obtain the 3D interest point response function value $\rho$ for every vertex $v$ on a 3D mesh model, those vertices with local maxima of final response function values are selected as the candidates of the 3D interest points. Specifically, we compare the value of response function \emph{$\rho$} for every vertex $v$ with those points in its neighborhood rings \emph{$\bm{V}_n(v),1\leq n \leq N$}. If the value $\rho$ of vertex $v$ is larger than all the values $\rho$ of those points in its neighborhood rings \emph{$\bm{V}_n(v),1\leq n \leq N$}, the vertex $v$ will be selected as the candidate of 3D interest point. Otherwise, vertex $v$ is not a 3D interest point. Fig. \ref{response_curves} also displays the final response function curve after non-maxima suppression of \emph{bird} 3D mesh model.

\subsection{Refinement of the Candidates of 3D Interest Points}
For candidates of 3D interest points of a 3D mesh model, although they have local maxima of final response function values $\rho$, there still exist some vertices with low value of response function value \emph{$\rho$}. Similar to the hard thresholding used in \cite{Blumensath2009265}, we utilize an $\ell_0$ norm based optimization method to refine the candidates of 3D interest points, where the vertices with low value of final response function value \emph{$\rho$} should be eliminated from the candidates.  Let \emph{$\bm{\rho} = [\rho(1),\rho(2),...,\rho(C)]$} be the set of response function values \emph{$\rho(j),j=1,2,...,C$} of 3D interest point candidates. We refine it using the following optimization model:
\begin{equation}
\begin{split}
\underset{\bm{x}} {\text{minimize}} \ \beta\lVert \bm{x}\lVert _0 + \lVert \emph{\bm{$\rho$}} - \emph{\bm{$\rho$}}\odot\bm{x}\rVert _2 ^2, \label{optimization_f}\\
\text{subject to} \ \ \ \ \ \ 
\bm{x} = [x_1,x_2,...,x_C]\\
x_j \in \{0, 1\},j=1,2,...,C\\
\end{split}
\end{equation}

\begin{figure}[tbp]
	\begin{minipage}[b]{1.0\linewidth}
		\centering
		\centerline{\includegraphics[width=0.6\linewidth]{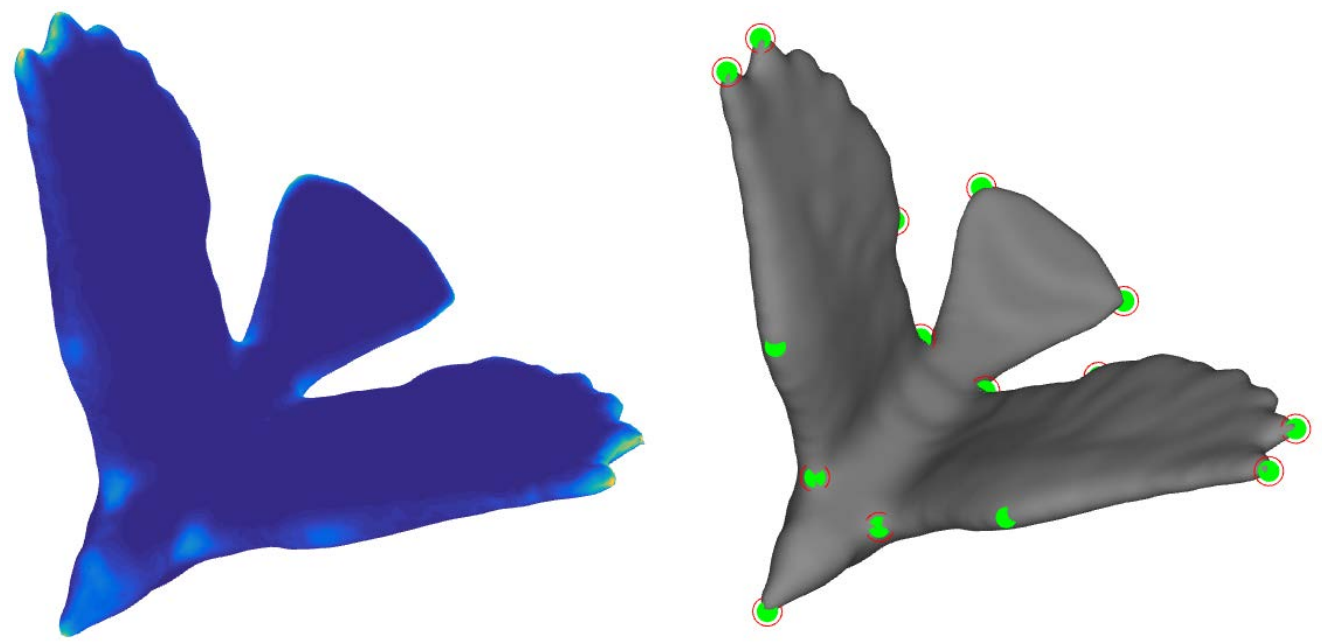}}
	\end{minipage}
	\caption{Left: final interest point response function map of \emph{bird} 3D mesh model; Right: candidate 3D interest points (green dots) and final 3D interest points (red circles).}
	\label{GMSR_bird_2}
\end{figure}

The first term of the target function (\ref{optimization_f}) is to constrain the number of 3D interest points.  \emph{$\beta$} is a penalty coefficient. The second term of the target function (\ref{optimization_f}) makes the \emph{$ \bm{\rho}_{opt} = \bm{\rho} \odot \bm{x} $} close to \emph{\bm{$\rho$}} as much as possible. \emph{$\bm{\rho}_{opt}$} is the Hadamard product of \emph{$\bm{x}$} and \emph{\bm{$\rho$}}. \emph{$\bm{x}$} is a vector with only two discrete elements in $x_j$ and it has the same dimension with \emph{\bm{$\rho$}}. \emph{$x_j=0$} indicates that the \emph{j}-th candidate of interest points has a low value of \emph{$\rho$} and vice versa. We utilize a greedy algorithm to solve the optimization model \cite{1337101}. Those points with \emph{$\rho_{opt}$} larger than zero are selected as the final 3D interest points. The right part of Fig. \ref{GMSR_bird_2} displays the candidates of 3D interest points (green dots) of \emph{bird} mesh model and its final 3D interest points (red circles) after sparse refinement respectively.

\section{Numerical Experiments}
\label{experiments}
In recent years, how to evaluate the performance of interest point detection algorithms has also been a popular research topic. Some famous evaluation benchmarks also demonstrate the performance of interest point detection algorithms to some extent, such as \cite{Compares_A_performance_evaluation_of_local_descriptors,Compares_Affine_Region_Detectors,Compares_evaluation_of_interest_point_detectors,Compares_interest_point_detector_and_feature_descriptor_survey,Compares_interesting_interest_point,APerformanceEvaluationofVolumetric3DInterestPointDetectors,3D_evaluation_repeatability_Jour,Acomparativeevaluationof3DkeypointdetectorsinaRGBDObjectDataset,SHREC2011,3D_evaluation_groundtruth}.
For 3D interest point detection, the evaluation benchmarks can be divided into two parts. The first one \cite{APerformanceEvaluationofVolumetric3DInterestPointDetectors,3D_evaluation_repeatability_Jour,Acomparativeevaluationof3DkeypointdetectorsinaRGBDObjectDataset,SHREC2011} usually measures the repeatability rate according to varying factors, which are usually designed for some special tasks like shape retrieval \cite{SHREC2011}. The other one \cite{3D_evaluation_groundtruth}\footnote{http://www.itl.nist.gov/iad/vug/sharp/benchmark/3DInterestPoint/} utilizes the  human generated ground truth data to evaluate the performance of 3D interest point detection algorithms. In this paper, we only focus on the problem of 3D interest point detection. So we evaluate the performance of our proposed GMSR based interest point detector and compare it with seven state-of-the-art 3D interest point detection algorithms - namely 3D Harris \cite{3D_Harris_Jour}, HKS \cite{3D_HKS}, Salient Points \cite{3D_Salient_point}, Mesh Saliency \cite{3D_Mesh_saliency_Jour} and Scale Dependent Corners \cite{3D_SD_corner}, 3D-CTARA \cite{3DCTARA}, 3D-ACRA \cite{3DCTARA} as well as our previous work \cite{3D_GMSR_Pre} under benchmark \cite{3D_evaluation_groundtruth}. The benchmark also provided results of tests on five algorithms \cite{3D_Harris_Jour,3D_Salient_point,3D_SD_corner,3D_Mesh_saliency_Jour,3D_HKS}.

\begin{figure*}[tbp]
	\begin{minipage}[b]{1.0\linewidth}
		\centering
		\centerline{\includegraphics[width=\linewidth]{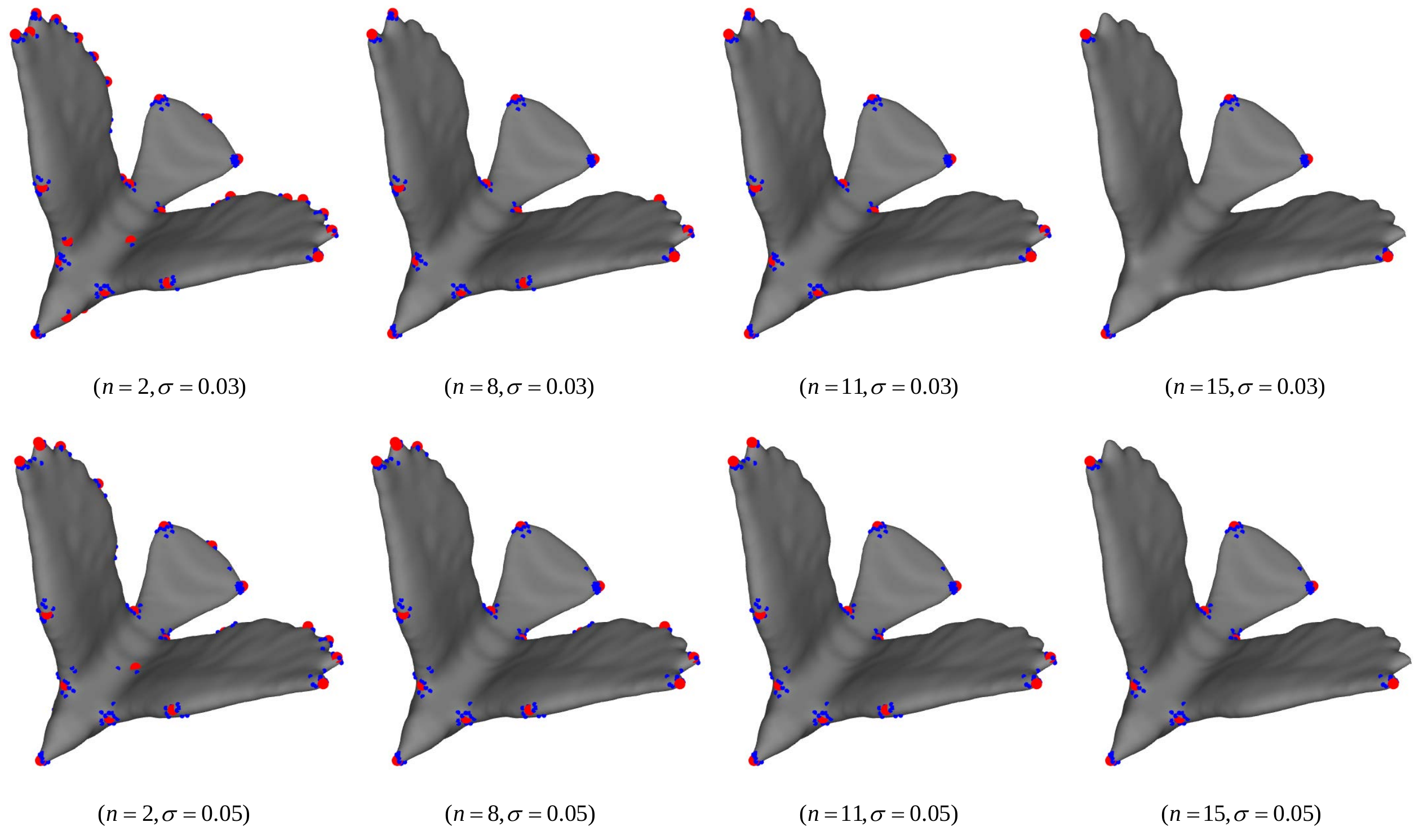}}
	\end{minipage}
	\caption{3D interest points of ground truth (red dots) of \emph{bird} mesh model marked by different human subjects $n$ (blue dots) and radius $\sigma$.}
	\label{bird2_gt}
\end{figure*}
\begin{figure*}[htbp]
	\begin{minipage}[b]{1.0\linewidth}
		\centering
		\centerline{\includegraphics[width= 1\linewidth]{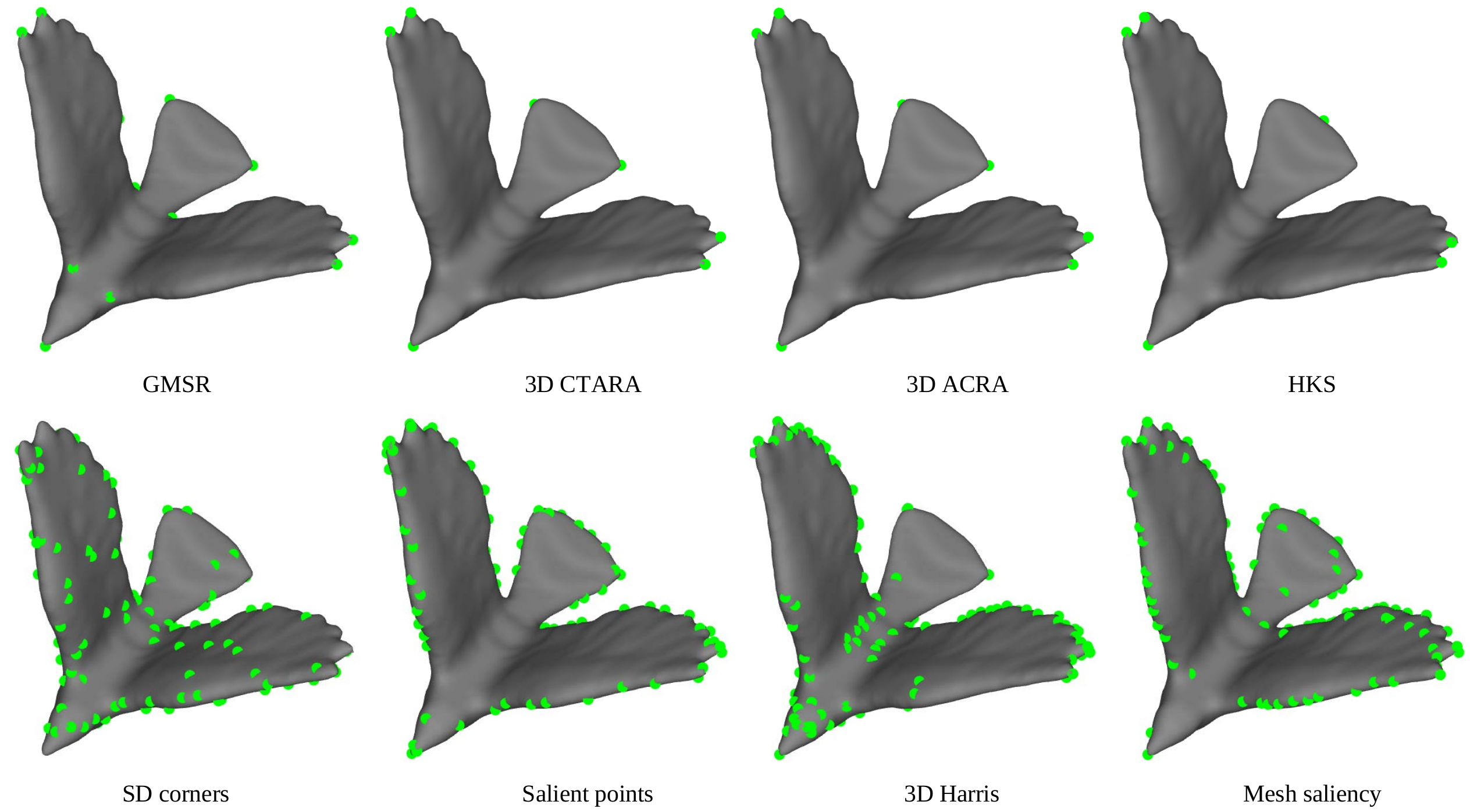}}
	\end{minipage}
	\caption{3D interest points of \emph{bird} mesh model detected by eight methods.}
	\label{bird_2}
\end{figure*}

\subsection{Datasets and Evaluation Metrics}
\subsubsection{Datasets}
We utilize the same benchmark datasets as those in \cite{3D_evaluation_groundtruth,3D_Song_spectral,3D_RF_keypoint}, where a web-based application is developed and utilized to collect ground truth of interest points on 43 3D mesh models. These mesh models are organized in two datasets. The first one (Dataset A) is constituted by 24 triangular mesh models and annotated by 23 human subjects. The other one (Dataset B) is constituted by 43 triangular mesh models and annotated by 16 human subjects. There are two criteria while constructing the ground truth: the radius of an interest region denoted as \emph{$\sigma$} and the number of human subjects \emph{$n$} that annotated a point within that interest region. For a fixed \emph{$\sigma$}, fewer ground truth points are observed as \emph{$n$} increases since a higher consensus among human subjects is needed. For a fixed \emph{$n$}, more ground truth points are observed as \emph{$\sigma$} increases, since we accept more variation on localization of the points annotated by the human subjects. Fig. \ref{bird2_gt} displays the 3D interest points of ground truth (red dots) marked by different human subjects $n$ (blue dots) and radius $\sigma$.

\subsubsection{Evaluation Metrics}
In benchmark \cite{3D_evaluation_groundtruth}, three evaluation metrics - namely False Positive Error (FPE), False Negative Error (FNE) and Weighted Miss Error (WME) are utilized to evaluate the performance of 3D interest point detection algorithms. But FPE and FNE can be misleading in isolation as discussed in \cite{3D_RF_keypoint}. So, Teran \cite{3D_RF_keypoint} adopted the Intersection Over Union (IOU) \cite{3D_IOU_scores} which combines False Positive (FP), False Negative (FN) and True Positive (TP) together as their main metric to evaluate the performance of the 3D interest point algorithms. In this paper, we also adopt the IOU metric as the main evaluation metric to evaluate the performance of our proposed GMSR 3D interest point algorithm and other seven state-of-the-art methods as well as our previous work \cite{3D_GMSR_Pre}. As mentioned before,	from another viewpoint, we can handle 3D interest point detection as a binary classification problem. So, in this paper, we also give the average $F_1$ score to further demonstrate the superiority of our proposed GMSR interest point detection algorithm.

Let \emph{$\bm{G}_M(n, \sigma)$} be the set of ground truth points and \emph{$\bm{A}_M$} be the 3D interest points detected by an algorithm on the 3D mesh model \emph{$\bm{M}(x,y,z)$}. For an interest point \emph{$g$} in set \emph{$\bm{G}_M(n, \sigma)$}, a geodesic neighborhood of radius \emph{$r$} is defined as \emph{$ \varDelta_r(g) = \{ p \in \bm{M}(x,y,z) | d(g,p) \leq r\} $} where \emph{$d(g,p)$} is the geodesic distance between points \emph{$g$} and \emph{$p$}. The parameter \emph{$r$} controls the localization error tolerance. A ground truth point \emph{$g$} is correctly identified if there exists a detected point \emph{$a \in \bm{A}_M$} in \emph{$\varDelta_r(g)$} and no other point in \emph{$\bm{G}_M(n,\sigma)$} closer to \emph{$a$}. IOU score at localization error tolerance \emph{$r$} can be calculated by
\begin{equation}
\text{IOU}(r) = \frac{TP}{FN+FP+TP},
\end{equation}
where \emph{$FP=N_A-N_C$} is the number of false positives and \emph{$FN=N_G-N_C$} represents the number of false negatives. \emph{$TP=N_C$} is the true positives. \emph{$N_G$} is the number of ground truth points, \emph{$N_C$} is the number of correctly detected points and \emph{$N_A$} denotes the number of detected interest points by the algorithm.

\subsection{Experimental Results}
\subsubsection{Relative Parameters Setting}
In this paper, there are some parameters which have influence on the performance of our proposed GMSR based detector. 

\begin{figure}[tbp]
	\begin{minipage}[b]{1.0\linewidth}
		\centering
		\centerline{\includegraphics[width= 0.5\linewidth]{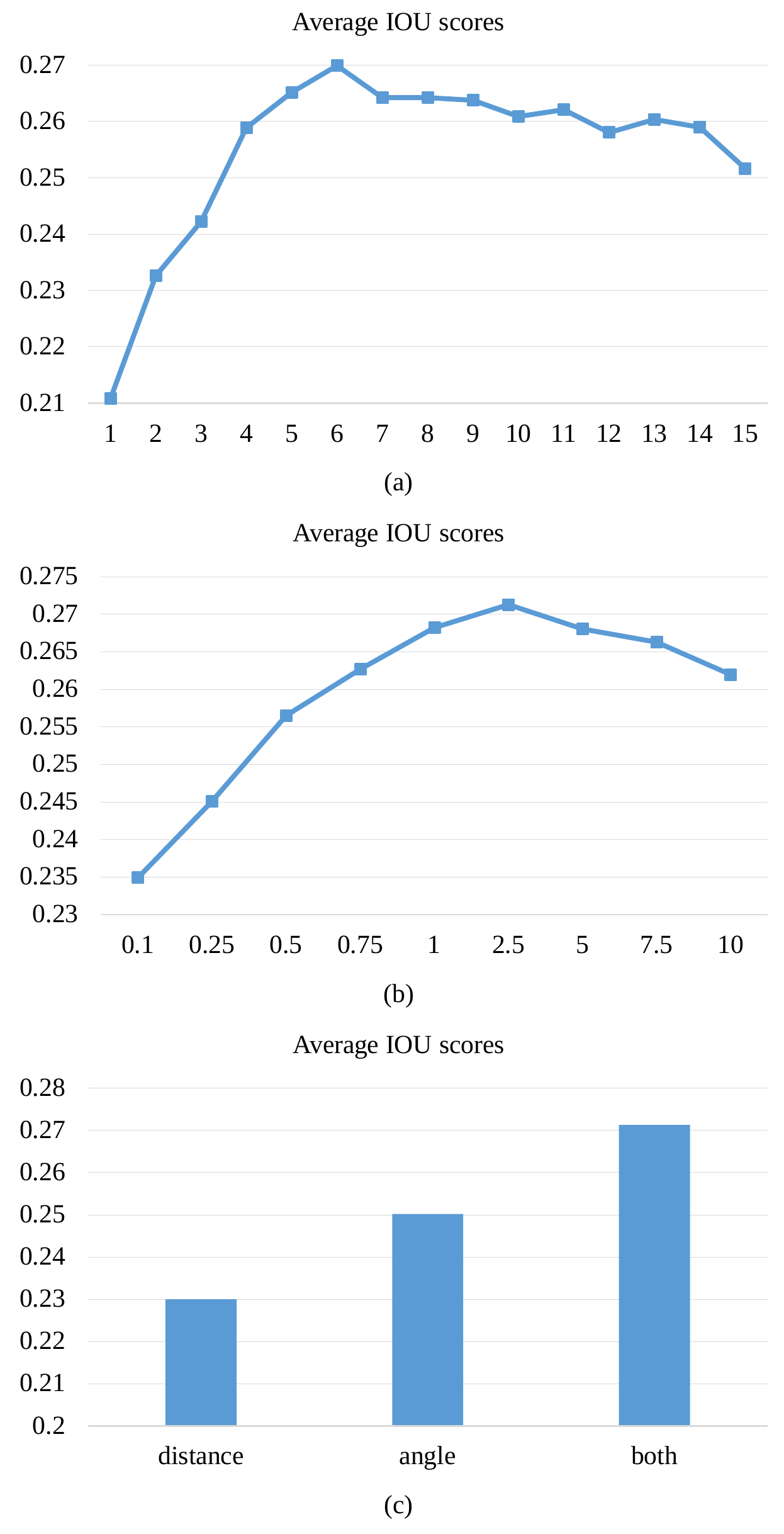}}
	\end{minipage}
	\caption{(a) Average IOU scores in terms of different neighborhood rings $K$; (b) Average IOU scores in terms of parameter $\alpha$. (c) Average IOU scores for using distance only, angle only and both of them.}
	\label{params}
\end{figure}

(a) Parameter $K$ in Section \ref{two_geometric_properties} has a direct influence on how many neighborhood vertices are utilized in our approach. Because interest point is a local feature, a small parameter $K$ will lead to the lack of enough available information to detect interest points. But a large parameter $K$ will damage the local structure information of current vertex. Fig. \ref{params} (a) shows the average IOU scores on benchmark \cite{3D_evaluation_groundtruth} in terms of rings $K$ in different neighborhood zones. In our approach, we set the parameter $K$ as 6. 

(b) Parameter $\alpha$ in (\ref{linera_norm}) in a weighting factor for two kinds of measures. Fig. \ref{params} (b) displays average IOU scores on benchmark \cite{3D_evaluation_groundtruth} in terms of different parameter $\alpha$. As shown in Fig. \ref{params} (c), compared with only one measure, we can see that the combination of two measures using a linear function could greatly improve the performance of GMSR based detector. We set parameter $\alpha$ as 2.5. 

(c) Parameter $N$ in Section \ref{non-maxima_supression}  and parameter $\beta$ in (\ref{optimization_f}) control the number and strength of detected interest points respectively. Parameter $N$ is directly related to the numbers of candidates of detected interest points. The smaller parameter $N$ is, the more candidates of interest points will be detected and vice versa. Parameter $\beta$ in (\ref{optimization_f}) is a penalty coefficient, which controls the number and strength of final interest points. The larger parameter $\beta$ is, the smaller the numbers of final interest points will be, and the higher the response function values of the final interest points will be. In  all the experiments, we set parameter $N$ as 10 and parameter $\beta$ as 0.03.

\subsubsection{Experiment Settings and Result Analysis}
In this section, we display the performance results of seven algorithms on Dataset A and Dataset B in terms of IOU scores and $F_1$ scores. It is important to note that the 3D interest points detected by these 3D interest point detection algorithms are constant when parameters $n/\sigma$ varies, but the ground truth is variable when parameters $n/\sigma$ varies according to \cite{3D_evaluation_groundtruth}.

Fig. \ref{GMSR_bird_2} displays the final response function map, candidates of 3D interest points (green dots) and final 3D interest points (red circles) of \emph{bird} mesh model detected by our proposed GMSR based detector. Visualizations of comparative results can be found in Fig. \ref{bird_2}, where 3D interest points of \emph{bird} detected by eight algorithms are displayed. We can see that the detection accuracy of our algorithm is obviously better than those of other methods, which means the 3D interest points detected by our approach are more in accord with human visual characteristics.

\begin{figure*}[htbp]
	\begin{minipage}[b]{1.0\linewidth}
		\centering
		\centerline{\includegraphics[width=0.8\linewidth]{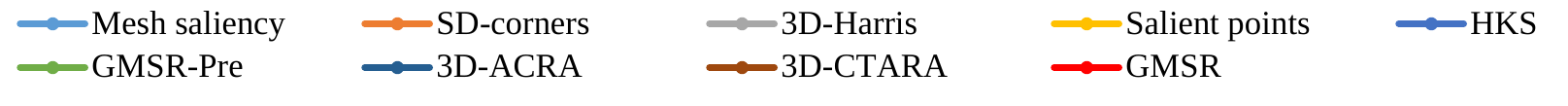}}
		\centerline{\includegraphics[width= \linewidth]{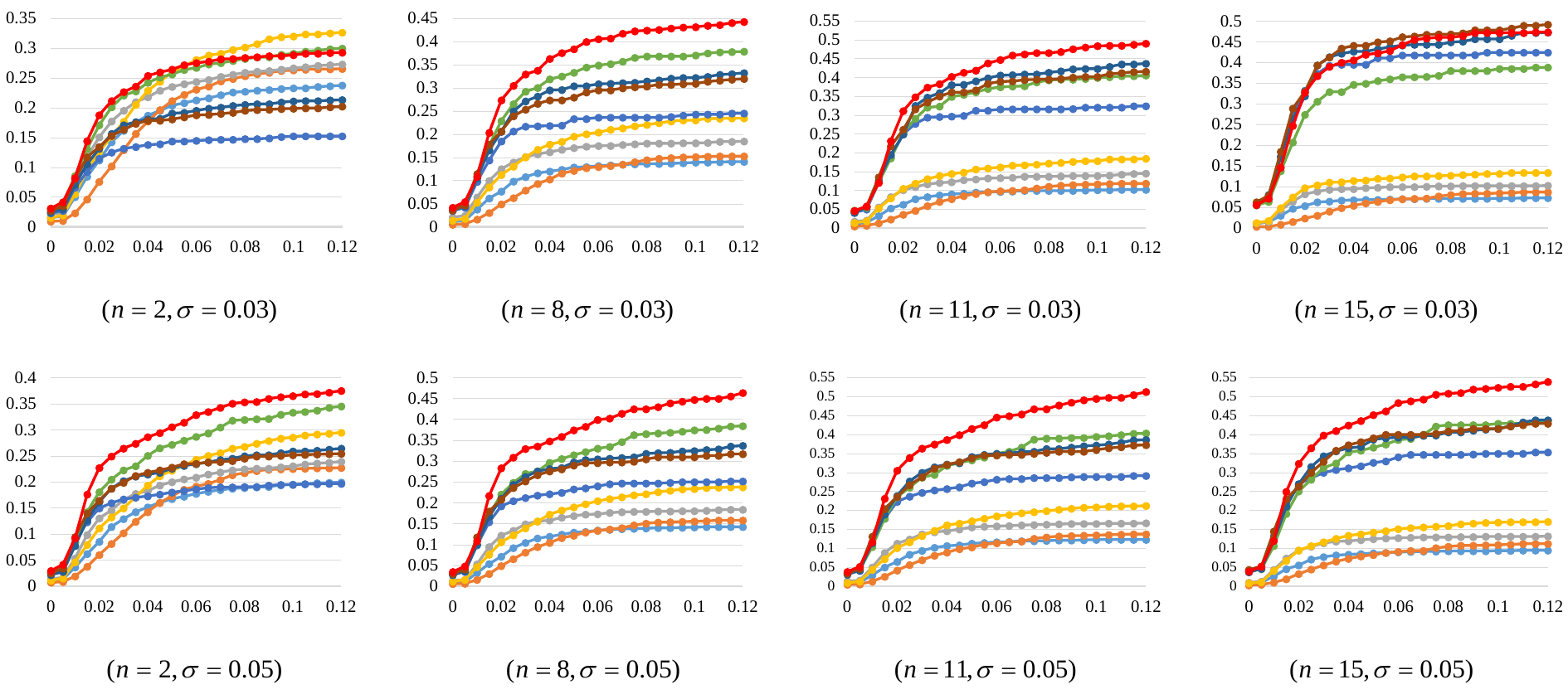}}
	\end{minipage}
	\caption{IOU scores for Dataset A at various parameter pairs $n/\sigma$ for different localization error tolerance $r$.}
	\label{details_iou_scores_A}
\end{figure*}
\begin{figure*}[htbp]
	\begin{minipage}[b]{1.0\linewidth}
		\centering
		\centerline{\includegraphics[width=0.8 \linewidth]{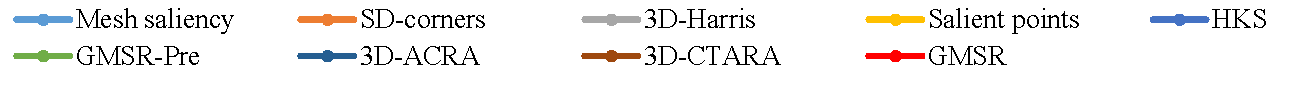}}
		\centerline{\includegraphics[width= \linewidth]{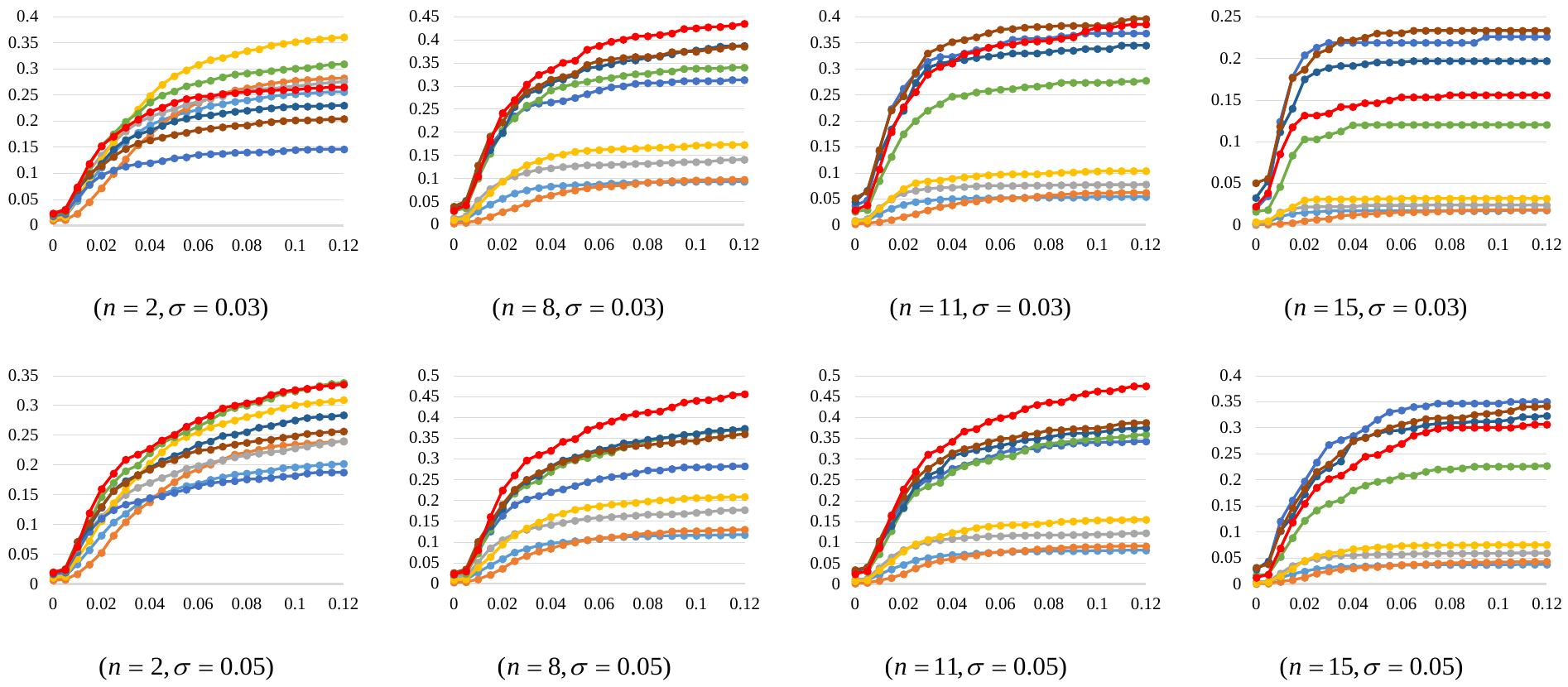}}
	\end{minipage}
	\caption{IOU scores for Dataset B at various parameter pairs $n/\sigma$ for different localization error tolerance $r$.}
	\label{details_iou_scores_B}
\end{figure*}
\begin{table*}[tbp]  
	\centering
	\renewcommand\arraystretch{1.2}
	\caption{Average IOU and $F_1$ scores on Dataset A (\emph{$n \in \{2,3,...,23\}, \sigma \in \{0.01,0.02,...,0.1\}, r \in \{0.005,0.01,...,0.12\}$}) and Dataset B (\emph{$n \in \{2,3,...,16\}, \sigma \in \{0.01,0.02,...,0.1\},r \in \{0.005,0.01,...,0.12\}$}) for eight methods.}
	\begin{tabular}{ccccccccccc}  % {llll} ±íÊ¾¸÷ÁÐÔªËØ¶ÔÆë·½Ê½£¬left-l,right-r,center-c
		\hline
		\cline{1-11}
		& &Mesh saliency  &SD-corners  &3D Harris  &Salient points &HKS &GMSR-Pre &3D ACRA &3D CTARA &GMSR\\ \hline     
		Dataset A &IOU &0.0783 &0.0758 &0.1044 &0.1245 &0.2470 &0.2658 &0.2799 &0.2854 &\textbf{0.3216} \\
		Dataset A &$F_1$ &0.1364 &0.1318 &0.1778 &0.2061 &0.3512 &0.3730 &0.3904 &0.3955 &\textbf{0.4334} \\
		
		Dataset B &IOU &0.0748 &0.0727 &0.1023 &0.1237 &0.2153 &0.2254 &0.2460 &0.2483 &\textbf{0.2706} \\
		Dataset B &$F_1$ &0.1311 &0.1273 &0.1737 &0.2037 &0.3159 &0.3273 &0.3541 &0.3552 &\textbf{0.3779} \\
		\hline
		\cline{1-11}
	\end{tabular}
	\label{overall_performance}
\end{table*}

In order to quantitatively analyze the performance of seven algorithms, we first test the IOU scores for Dataset A and Dataset B at various parameter $n/\sigma$ pairs. Fig. \ref{details_iou_scores_A} and Fig. \ref{details_iou_scores_B} show the IOU scores at several parameters $n/\sigma$ pairs for Dataset A and Dataset B respectively, in which we set $n\in \{2,8,11,15\}$, $\sigma \in \{0.03,0.05\}$ and $r \in \{0.005,0.01,0.015,...,0.12\}$. From Fig. \ref{details_iou_scores_A}, we can see that our proposed GMSR based 3D interest point detector performs best in all the situations except for one situation in Dataset A where \emph{$n=2,\sigma=0.03$}. When localization error tolerance \emph{$r$} is relatively large, the superiority of our proposed GMSR based 3D interest point detection algorithm is more significant. For Dataset B, when parameter $n$  is large, e.g. $n = 15$, HKS \cite{3D_HKS} outperforms our proposed GMSR based approach, because the HKS method tends to detect a very small number of very salient points. But for others parameter pairs $n/\sigma$, like $n = 8/\sigma=0.03,n=8/\sigma=0.05,n=11/\sigma=0.05$ etc, our proposed GMSR based approach performs best. As shown in Fig. \ref{details_iou_scores_A} and Fig. \ref{details_iou_scores_B}, we can see that HKS \cite{3D_HKS} algorithm tends to detect a very small number of strongly salient points but can't detect even slightly more ambiguous interest points, while our proposed GMSR based approach tends to detect these points with a higher consensus among human subjects as well as those ambiguous interest points.

\begin{figure}[htbp]
	\begin{minipage}[b]{1.0\linewidth}
		\centering
		\centerline{\includegraphics[width= 0.5\linewidth]{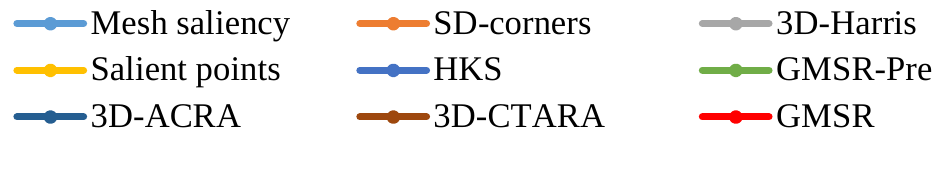}}
		\centerline{\includegraphics[width=0.6 \linewidth]{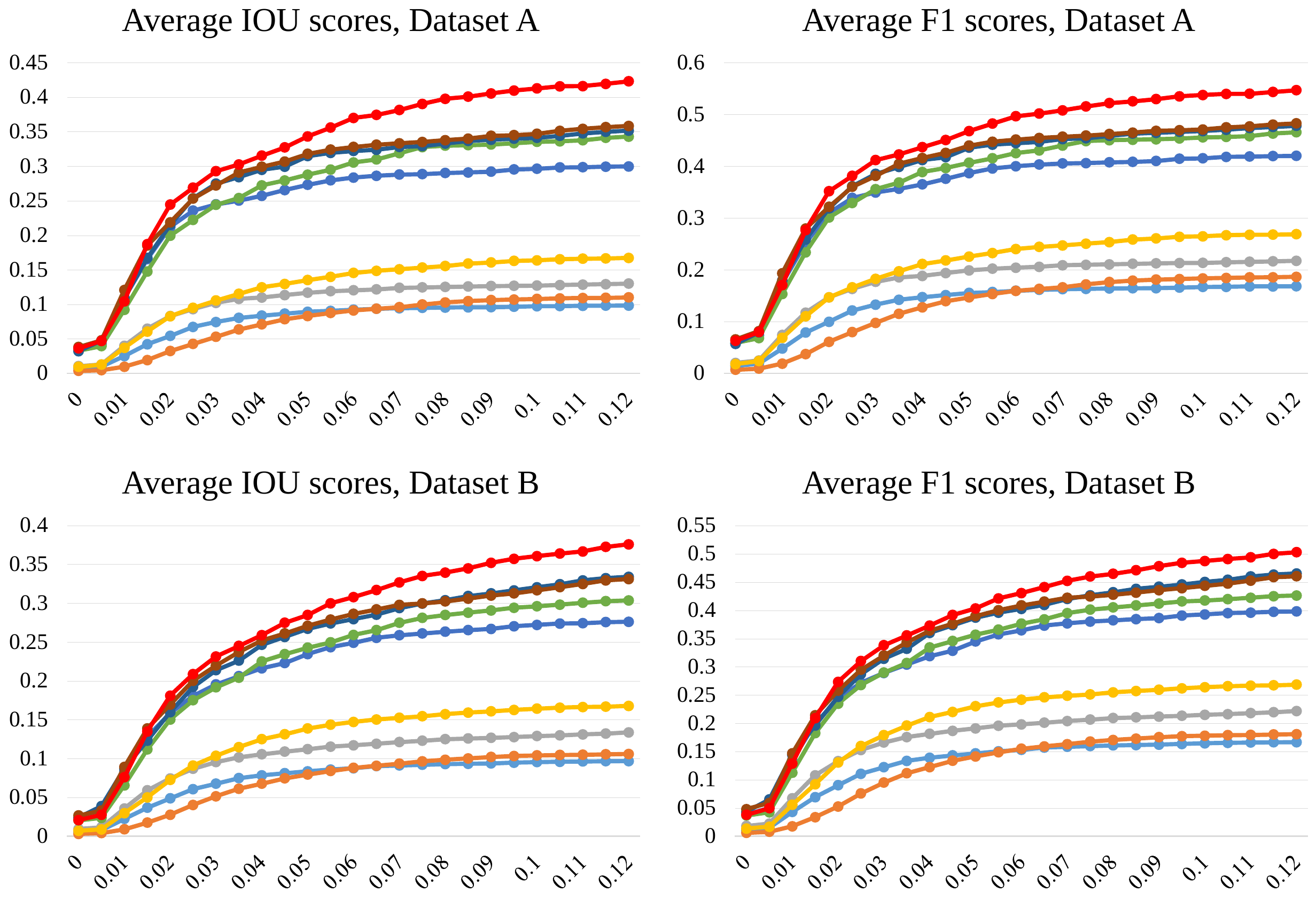}}
	\end{minipage}
	\caption{Average IOU scores and $F_1$ scores for Dataset A ($n \in \{2,3,...,23\}, \sigma \in \{0.01,0.02,...,0.1\}$) and Dataset B ($n \in \{2,3,...,16\}, \sigma \in \{0.01,0.02,...,0.1\}$) at different localization error tolerance $r$.}
	\label{average_iou_f1}
\end{figure}

To reach an overall performance ranking, we calculate the average IOU scores and $F_1$ scores over all the parameter pairs \emph{$n/\sigma$} settings in Dataset A  and Dataset B. Fig. \ref{average_iou_f1}  shows the average IOU scores and $F_1$ scores for Dataset A (\emph{$n \in \{2,3,...,23\}, \sigma \in \{0.01,0.02,...,0.1\}$}) and  Dataset B (\emph{$n \in \{2,3,...,16\}, \sigma \in \{0.01,0.02,...,0.1\}$}) at different localization error tolerance $r$. In all the localization error tolerance $r$, our proposed GMSR based approach outperforms others algorithms in terms of both IOU scores and $F_1$ scores, especially when localization error tolerance $r$ is relatively large. Besides, TABLE \ref{overall_performance} displays the average IOU scores and $F_1$ scores of Dataset A and Dataset B, where $n \in \{2,3,...,23\}, \sigma \in \{0.01,0.02,...,0.1\}, r \in \{0.005,0.01,...,0.12\}$ for Dataset A and $n \in \{2,3,...,16\}, \sigma \in \{0.01,0.02,...,0.1\}, r \in \{0.005,0.01,...,0.12\}$ for Dataset B. We can see that our proposed GMSR based 3D interest point detector performs best in terms of both IOU evaluation metric and $F_1$ score metric.

\section{Conclusion}
In this paper, we have proposed a new method for detecting robust 3D interest points of 3D mesh models based on geometric measures and sparse refinement. Two intuitive and effective geometric measures are defined in multi-scale space to distinguish 3D interest points from edges and flat areas effectively. We combine the two geometric measures via a linear function to represent their relationship approximately. For any vertex $v$, its interest point response function values at different scales are multiplied as the final one, which can effectively improve the 3D saliency measure of true 3D interest points and suppress the 3D saliency measure of pseudo 3D interest points. Those vertices with local maxima of final response function values are selected as the candidates of 3D interest points. Finally, an $\ell_0$ norm based optimization method is used to refine 3D interest points by constraining the number of 3D interest points. Experimental results demonstrate that our proposed GMSR based 3D interest point detector performs best in terms of both IOU scores metric and $F_1$ score metric as compared with previously proposed algorithms. The interest points detected by our approach are more in accord with human visual characteristics.

\bibliographystyle{IEEEtran}
\bibliography{citation}


%\vfill


\end{document}